\documentclass[times,twocolumn,final,authoryear]{elsarticle}

\usepackage{prletters}
\usepackage[ruled,vlined]{algorithm2e}

\journal{Pattern Recognition Letters}

\usepackage{url}
\usepackage{graphicx}
\usepackage{amssymb}
\usepackage{amsthm}
\usepackage{amsmath}
\usepackage{multirow}
\usepackage{booktabs}

\begin{document}

\begin{frontmatter}


\title{Mixing Up Contrastive Learning: \\ Self-Supervised Representation Learning for Time Series}




\author[label1]{Kristoffer Wickstrøm}
\ead{kristoffer.k.wickstrom@uit.no}

\address[label1]{UiT Machine Learning Group at the Dept. of Physics and Technology, UiT the Arctic University of Norway, Tromsø NO-9037, Norway}
\address[label2]{Norwegian Computing Center, Dept. SAMBA, P.O. Box 114 Blindern,
NO-0314 Oslo, Norway}
\address[label3]{Dept. of Gastrointestinal Surgery, University Hospital of North Norway (UNN), Tromsø, Norway}

\author[label1,label2]{Michael Kampffmeyer}
\author[label1,label3]{Karl Øyvind Mikalsen}
\author[label1,label2]{Robert Jenssen}

\begin{abstract}
The lack of labeled data is a key challenge for learning useful representation from time series data. However, an unsupervised representation framework that is capable of producing high quality representations could be of great value. It is key to enabling transfer learning, which is especially beneficial for medical applications, where there is an abundance of data but labeling is costly and time consuming. We propose an unsupervised contrastive learning framework that is motivated from the perspective of label smoothing. The proposed approach uses a novel contrastive loss that naturally exploits a data augmentation scheme in which new samples are generated by mixing two data samples with a mixing component. The task in the proposed framework is to predict the mixing component, which is utilized as soft targets in the loss function. Experiments demonstrate the framework's superior performance compared to other representation learning approaches on both univariate and multivariate time series and illustrate its benefits for transfer learning for clinical time series.
\end{abstract}


\end{frontmatter}


\section{Introduction}\label{S:1}
\noindent Learning a useful representation of time series without labels is a challenging task. Nevertheless, time series are a typical data type in numerous domains where the lack of labeled data is a common challenge. Particularly in the medical domain there can often be an abundance of data but labeling can be costly and challenging \citep{missdata}. Learning useful representations from unlabeled data would be of great benefit in such scenarios. In particular, it could enable transfer learning for clinical time series. Transfer learning is the practice of transferring knowledge from a source domain to a target domain \citep{5288526}. Such a technique enables researchers to exploit large unlabeled datasets to train more robust and precise systems on small labeled datasets.

Learning useful representations is an active area of research in machine learning \citep{Bengio2013RepresentationLA, 9086055}, with encouraging results in recent works on image representation learning \citep{Chen2020ASF, MoCocvpr, ByolGrill}. Many of such recent works have used contrastive learning for learning useful features, and these works exploits prior information about noise invariances in the image data. However, time series data constitute a highly heterogeneous data source, and invariances can differ completely between different datasets.

Contrastive learning is a type of self-supervised representation learning where the task is to discriminate between different views of the sample, where the different views are created through data augmentation that exploit prior information about the structure in the data. Data augmentation is typically performed by injecting noise into the data. Recent advances in contrastive learning have been particularly prominent for image data, as there exists a wide range of applicable augmentation schemes \citep{Shorten2019ASO, Chen2020ASF} that are suitable for natural images. On the other hand, data augmentation for time series based on the injection of noise can be more challenging because of the heterogeneous nature of time series data and the lack of generally applicable augmentations.

This paper introduces a novel self-supervised learning framework that naturally exploits a recent data augmentation scheme called mixup \citep{zhang2018mixup}. The mixup data augmentation scheme creates an augmented sample through a convex combination of two data points and a mixing component. Such an approach allows for a natural generation of new data points, as augmented samples are generated through a combination of samples from the data distribution. In the proposed framework, the task is to predict the strength of the mixing component based on the two data points and the augmented sample, which is motivated by recent research on label smoothing \citep{labelS}. Label smoothing refers to the concept of adding noise to the labels, such that the targets are no longer hard 0 and 1 targets, but soft targets in the range between 0 and 1. This has been shown to increase performance and reduce overconfidence in deep learning-based approaches \citep{labelS}. The proposed framework shows encouraging results when evaluated on the UCR \citep{UTS} and UEA \citep{MTS} databases and compared to a number of baselines. Furthermore, we show how the proposed method can be used to enable transfer learning for clinical time series. Experiments illustrate that self-supervised pre-training can increase both performance and convergence speed for deep learning-based classification of clinical time series.

Our contributions are:

\begin{enumerate}
    \item A novel contrastive learning framework that is motivated through the concept of label smoothing and is based on predicting the amount of mixing between data points.
    \item An extensive evaluation of the proposed method with comparison to a number of baselines.
    \item We show how the proposed method enables transfer learning clinical time series, which leads to an increase in performance when classifying echocardiograms.
\end{enumerate}

\section{Mixup Contrastive Learning}

We outline the proposed framework for contrastive representation learning of time series. We propose a new contrastive loss that naturally exploits the information from the data augmentation procedure. Before we present our new contrastive learning framework, we introduce some notation. Our presentation will be based on univariate time series (UTS), but is also extended to multivariate time series (MTS) in the experiments. Let a UTS, $x$, be defined as a sequence of real numbers ordered in time, $x=\{x(t)\in\mathbb{R}|t=1,2,\cdots,T\}$, where $t$ denotes each time step and $T$ denotes the length of the UTS. Vectorial data will be denoted in lowercase bold $\mathbf{x}$.

A common approach to contrastive learning is to use a neural network-based encoder to transform the data into a new representation \citep{Chen2020ASF}. The encoder is trained by passing different augmentations of the same sample through the encoder and a projection head, before applying a contrastive loss. The goal of contrastive learning is to embed similar samples in close proximity by exploiting the invariances in the data. After training, the task dependent projection head is discarded and the encoder is kept for down-stream tasks.

The data augmentation scheme used to create different views of the same sample is crucial for learning a useful representation. However, care must be taken when determining the set of transformations to apply. The potential invariances of time series are rarely known in advance, and incautious application can result in a representation where unalike samples are embedded in close proximity \citep{pmlrJeeheh}. For instance, a transformation like rotation that is common to apply for natural images can completely change the nature of a time series by changing the trend of the data.

In this work, we opt for a data augmentation scheme based on creating new samples through convex combinations of training examples referred to as mixup \cite{zhang2018mixup}. Given two time series $x_i$ and $x_j$ drawn randomly from our training data, an augmented training example can be constructed as follows:

\begin{equation}\label{eq:mixupEQ}
    \tilde{x} = \lambda x_i + (1-\lambda)x_j.
\end{equation}

\begin{figure}[t]
    \centering
    \includegraphics[width=0.975\columnwidth]{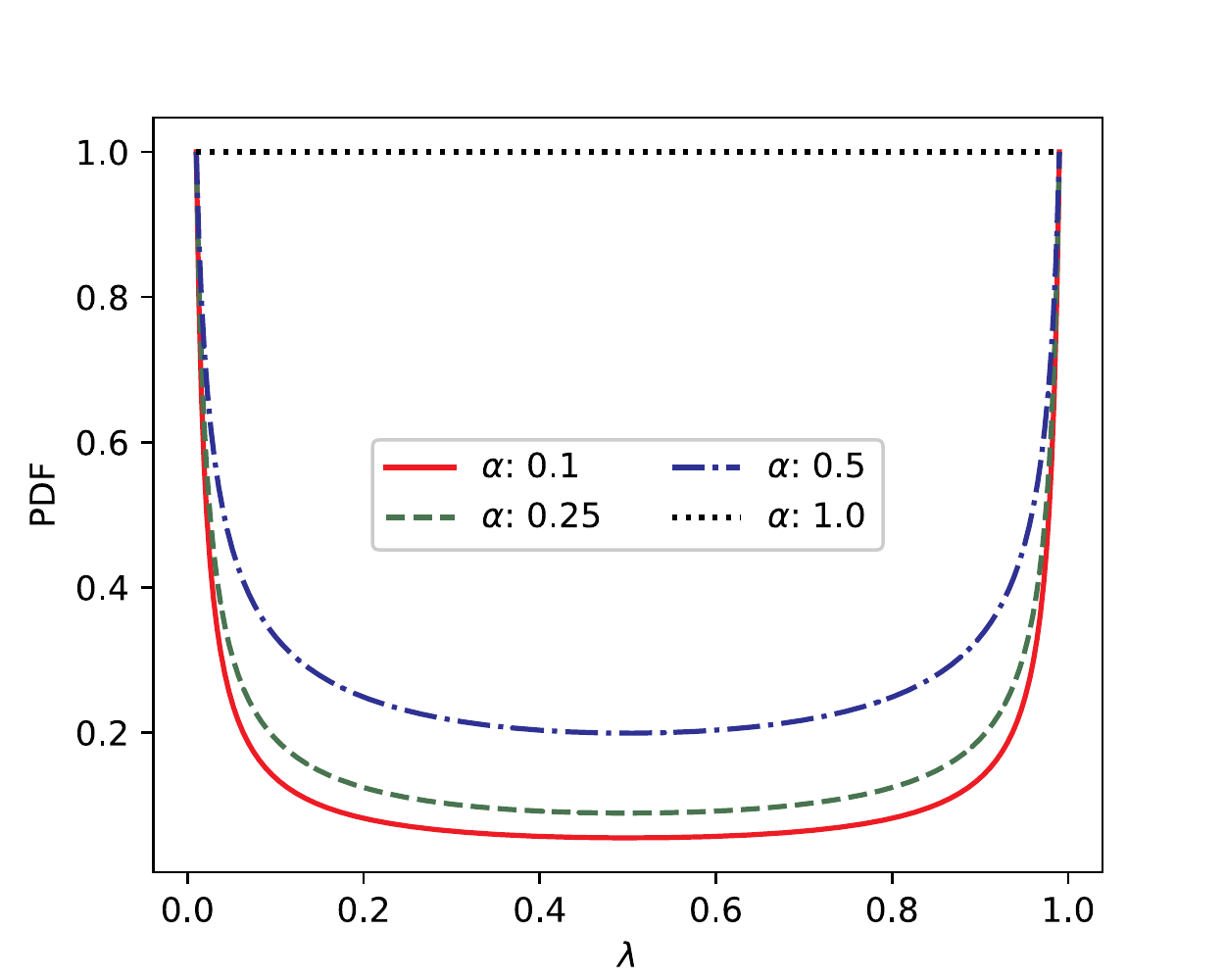}
    \caption{The beta distribution for different values of $\alpha$. As $\alpha$ approaches 1 the distribution tends towards a uniform distribution. Larger $\alpha$ results in more mixing of samples.}
    \label{fig:betaDist}
\end{figure}

Here, $\lambda\in[0,1]$ is a mixing parameter that determines the contribution of each time series in the new sample, where $\lambda \sim \textrm{Beta}(\alpha, \alpha)$ and $\alpha\in (0, \infty)$. The distribution of $\lambda$ for different values of $\alpha$ is illustrated in Figure \ref{fig:betaDist}. The choice of this augmentation scheme is motivated by avoiding the need to tune a noise parameter based on specific datasets but instead automatically generating data samples based on the specific dataset. Moreover, the information in the mixing parameter $\lambda$ can be exploited to produce a novel contrastive loss that is described in the following section. In a nutshell, the proposed framework is based on transforming the task from predicting hard 0 and 1 targets to soft targets $\lambda$ and $1-\lambda$. This is motivated by recent research on label smoothing that has shown how such regularization can lead to increased performance and less overconfidence in deep learning \citep{labelS}.

\subsection{A Novel Contrastive Loss for Unsupervised Representation Learning of Time Series}

\begin{figure}[t]
    \centering
    \includegraphics[width=0.975\columnwidth]{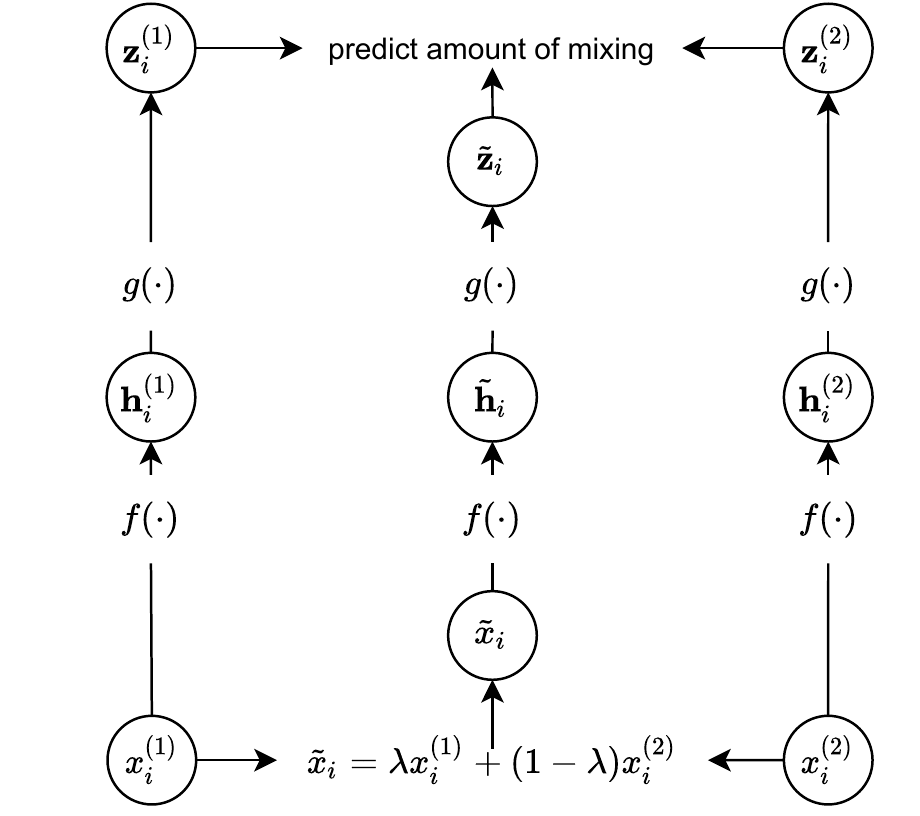}
    \caption{The proposed  framework. Two minibatches are sampled randomly from the data and combined using Equation \ref{eq:mixupEQ}. All samples are passed though an encoder $f(\cdot)$ resulting in a representation that can be used for down-stream tasks. Next, this representation is transformed using a projection head $g(\cdot)$ into a representation where the proposed contrastive loss is applied.}
    \label{fig:CLframework}
\end{figure}

We propose a new contrastive loss function that naturally exploits the information from the mixing parameter $\lambda$. At each training iteration, a new $\lambda$ is drawn randomly from a beta distribution, and two minibatches of size $N$, $\{x_1^{(1)}, \cdots, x_N^{(1)}\}$ and $\{x_1^{(2)}, \cdots, x_N^{(2)}\}$, are drawn randomly from the training data. Applying Equation \ref{eq:mixupEQ}, the two minibatches are used to create a new minibatch of augmented samples, $\{\tilde{x}_1, \cdots, \tilde{x}_N\}$. All three minibatches are passed through the encoder, $f(\cdot)$, that transforms the data into a new representation, $\{\mathbf{h}_1^{(1)}, \cdots, \mathbf{h}_N^{(1)}\}$, $\{\mathbf{h}_1^{(2)}, \cdots, \mathbf{h}_N^{(2)}\}$, and $\{\tilde{\mathbf{h}}_1, \cdots, \tilde{\mathbf{h}}_N\}$, which can be used for down-stream tasks. Next, the new representations are again transformed into a task-dependent representation, $\{\mathbf{z}_1^{(1)}, \cdots, \mathbf{z}_N^{(1)}\}$, $\{\mathbf{z}_1^{(2)}, \cdots, \mathbf{z}_N^{(2)}\}$, and $\{\tilde{\mathbf{z}}_1, \cdots, \tilde{\mathbf{z}}_N\}$, by the projection head, $g(\cdot)$, where the contrastive loss is applied. The framework is illustrated in Figure \ref{fig:CLframework}. Using this notation, our proposed contrastive loss for a single instance is applied on the representation produced by the projection head and is defined as:

\begin{align*}
    l_i &= -\lambda\text{log}\frac{\text{exp}(\frac{D_C(\tilde{\mathbf{z}}_i, \mathbf{z}_i^{(1)})}{\tau})}{\sum\limits_{k=1}^{N}\big(\text{exp}(\frac{D_C(\tilde{\mathbf{z}}_i, \mathbf{z}^{(1)}_k)}{\tau})+\text{exp}(\frac{D_C(\tilde{\mathbf{z}}_i, \mathbf{z}^{(2)}_k)}{\tau})\big)} \\
    &-(1-\lambda)\text{log}\frac{\text{exp}(\frac{D_C(\tilde{\mathbf{z}}_i, \mathbf{z}_i^{(2)})}{\tau})}{\sum\limits_{k=1}^{N}\big(\text{exp}(\frac{D_C(\tilde{\mathbf{z}}_i, \mathbf{z}^{(1)}_k)}{\tau})+\text{exp}(\frac{D_C(\tilde{\mathbf{z}}_i, \mathbf{z}^{(2)}_k)}{\tau})\big)},
\end{align*}

where $D_C(\cdot)$ denotes the cosine similarity and $\tau$ denotes a temperature parameter, as in recent works on contrastive learning \citep{Chen2020ASF}. The loss will be referred to as the MNT-Xent loss (the mixup normalized temperature-scaled cross entropy loss). The proposed loss changes the task from identifying the positive pair of samples, as in standard contrastive learning, to predicting the amount of mixing. Moreover, neural networks are known to be overly confident in predictions far from the training data \citep{Hein2019WhyRN}, but the proposed loss will discourage overconfidence since the model is tasked with predicting the mixing factor instead of a hard 0 or 1 decision.

\section{Experiments and Results}

We evaluate the proposed framework on an extensive number of both UTS and MTS datasets, and compare against well known baselines. Also, we demonstrate how the proposed methodology enables transfer learning in clinical time series.

\subsection{Evaluating Quality of Representation}

A common approach for evaluating the usefulness of an unsupervised contrastive learning framework is training a simple classifier on the learned representation \citep{8099559, Caron}. We use a 1-nearest-neighbor (1NN) classifier to evaluate quality of different representations, as suggested by \citet{UTS}. This is motivated by the simplicity of the 1NN classifier, which requires no training and minimal hyperparameter tuning. Furthermore, the 1NN classifier is highly dependent on the representation to achieve good performance and is therefore a good indicator of the quality of the learned representation. The proposed methodology, referred to as mixup contrastive learning (MCL), is evaluated on the UCR archive \citep{UTS}, which consists of 128 UTS datasets, and the UEA archive \citep{MTS}, which consists of 30 MTS datasets. We compare with several baselines that span different types of time series representations:

\begin{figure*}[htb]
    \centering
    \includegraphics[width=0.975\textwidth]{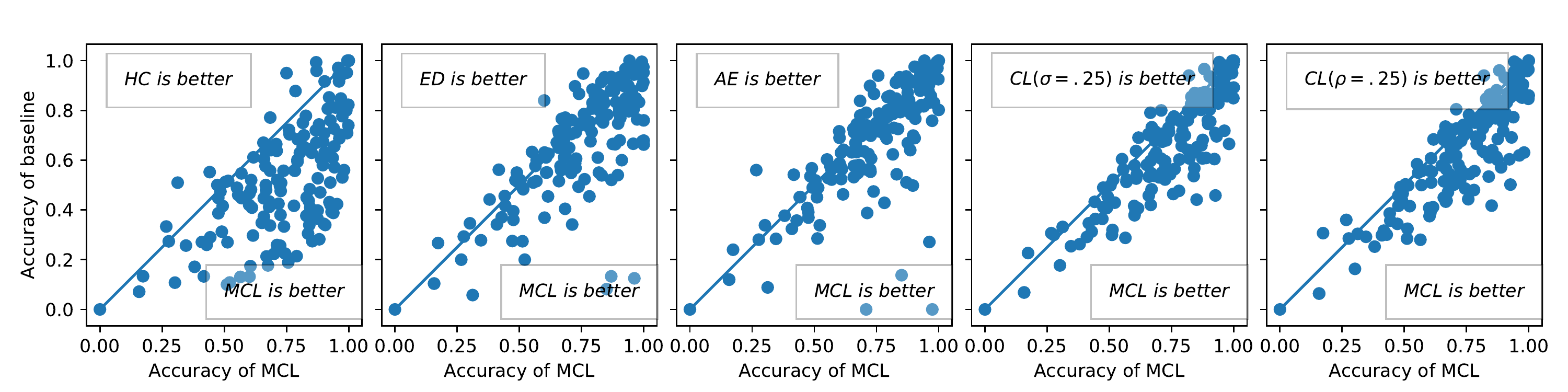}
    \caption{Accuracy on each dataset from the UCR and UEA databases. Each point represents the accuracy on one dataset, with the baseline along the vertical axis and the MCL along the horizontal axis. The diagonal line indicates where two methods have similar performance. Points above this line indicates that the baselines gives better performance and points below this line indicates that the MCL gives better performance. The figure shows that the proposed method provides superior performance to the baselines, as the majority of the points lie below the diagonal line.}
    \label{fig:scatter}
\end{figure*}

\begin{itemize}
    \item Handcrafted features (HC): Extract the maximum, minimum, variance and mean value of each time series. This is an elementary and well-known approach that will act as a simple baseline.
    \item Raw input features (ED): Using the raw time series as input without any alterations. This will demonstrate if it is beneficial to transform the time series.
    \item Autoencoder features (AE): A deep learning-based baseline using an autoencoder framework. We use the same network as with the proposed method, and a mirrored encoder for the decoder. The model is trained using a mean squared error reconstruction loss for 250 epochs. Autoencoder-based learning of features from time series with a reconstruction loss is a typical approach in the literature \citep{9349150, ijcaiex2}
    \item Contrastive learning features (CL): A deep learning-based baseline based on the widely used SimCLR framework \citep{Chen2020ASF}. We use the same network as with the proposed method, but with different data augmentation and the standard contrastive loss of \citet{Chen2020ASF} instead of the mixup contrastive loss. We consider two data augmentation techniques, gaussian noise with a variance of 0.25 (CL $(\sigma=0.25)$) and dropout noise with a dropout rate of 0.25 (CL $(\rho=0.25)$). These noise parameters represent an average amount of noise suitable for most datasets.
\end{itemize}

\begin{table}[htb]
\centering
\caption{Accuracy and ranking of a 1NN classifier on different representations averaged over all datasets. Results show that the representation obtained from the proposed method results in better performance across all metrics.}
\vspace{0.25cm}
\resizebox{\columnwidth}{!}{%
\begin{tabular}{@{}l||c|c|c|c@{}}
\toprule
           & \multicolumn{2}{c|}{UCR}             & \multicolumn{2}{c}{UEA}             \\ \midrule
Features & Avg accuracy & Avg ranking & Avg accuracy & Avg ranking \\ \midrule \midrule
HC         & .520        & 1.63       & .560        & 2.80       \\ \midrule
ED         & .686        & 3.86     & .585        & 3.56     \\ \midrule
AE         & .702        & 4.00     & .587        & 3.56     \\ \midrule
CL $(\sigma=.25)$        & .666        & 3.41      & .573        & 3.16     \\ \midrule
CL $(\rho=.25)$        & .660        & 2.99      & .570        & 3.06     \\ \midrule
MCL        & \textbf{.759}       & \textbf{4.81}      & \textbf{.627}        & \textbf{4.26}    \\ \bottomrule
\end{tabular}%
}
\label{tab:results}
\end{table}

\begin{figure}[htb]
    \centering
    \includegraphics[width=0.975\columnwidth]{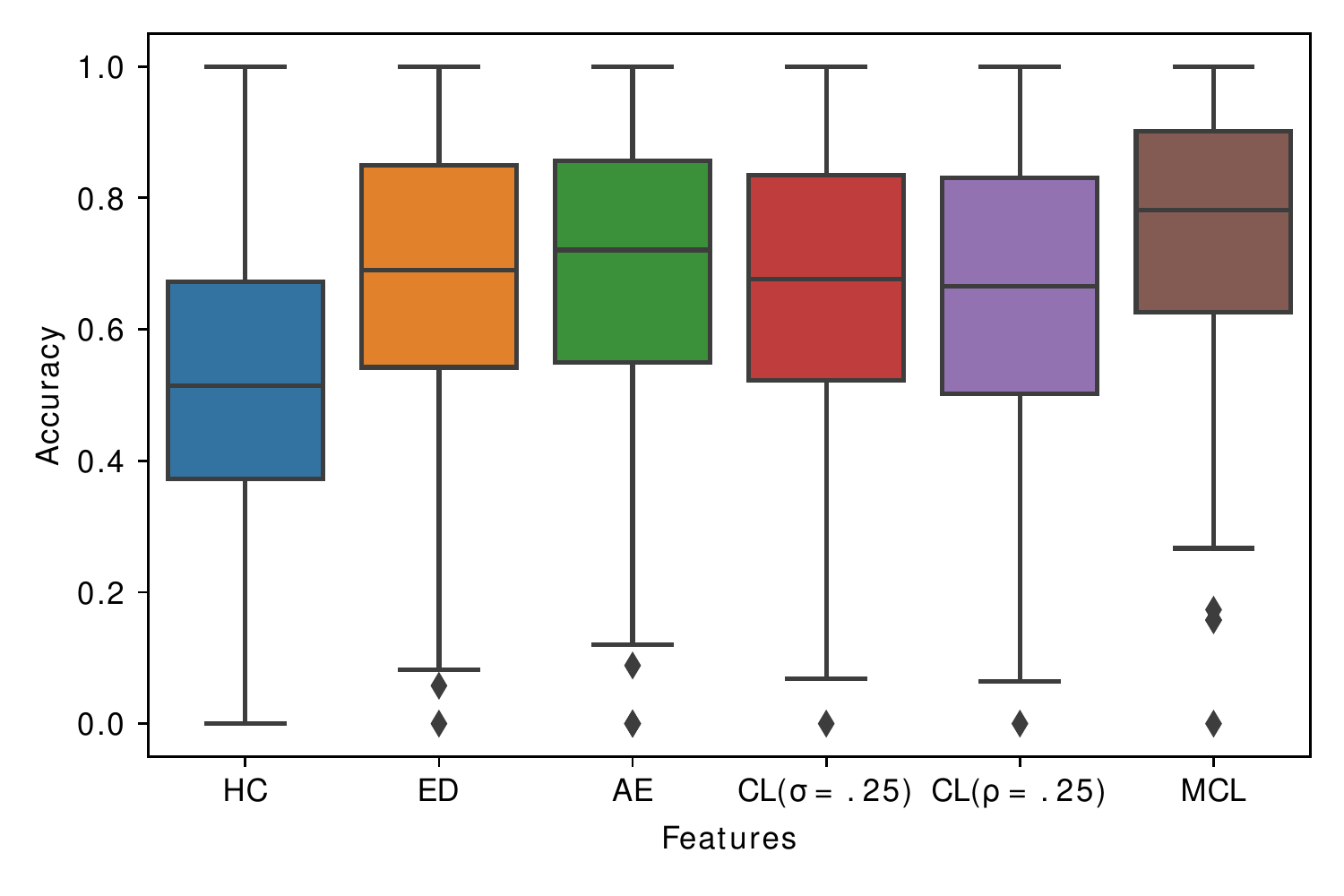}
    \caption{Boxplot of accuracy across all datasets of the UCR and UEA databases. Figure shows that the proposed framework produces representations that yield better performance at a more consistent rate. The whiskers indicated the lower and upper quartiles, with outliers indicated through black dots. }
    \label{fig:boxplot}
\end{figure}

We use the fully convolutional network (FCN) proposed by \citet{7966039} as an encoder $f$ for all contrastive learning approaches reported in this work. The FCN consists of three convolutional layers, each followed by batch normalization \citep{batchnorm} and a rectified linear unit activation function, and an adaptive average pooling layer. The convolutional layers consist of 128, 256, and 128 filters from first to third layer. This choice is motivated by the FCN's strong performance on a number of time series benchmark tasks (Wang et al. (2017)) and its simplicity. Specifically, the encoder representation will be the output of the average pooling layer. For the projection head $g$, we use a two-layer neural network with 128 neurons in each layer and separated by a rectified linear unit non-linearity, inspired by \citet{Chen2020ASF}. All models are optimized using the ADAM optimizer \citep{ADAM} for 1000 epochs, with the temperature parameter $\tau$ is set to 0.5 as suggested by \citet{Chen2020ASF}, and the $\alpha$ parameter set to 0.2 as suggested by \citet{zhang2018mixup}. Statistical significance is determined using a pairwise t-test, where bold numbers indicate significance at a significance level of 0.05. The accuracy and ranking of the learned features (AE, CL and MCL) are based on the average across 5 training runs at the last epoch.

Table \ref{tab:results} displays the results of the evaluation of the quality of the representation obtained through different representation learning approaches. Results on individual datasets are displayed in the Appendix. Table \ref{tab:results} shows that the simple HC baseline results in poor performance, even compared to no transformation of input (ED). Furthermore, the learned CL feature baseline gives comparable results to the ED features, while the AE features give a slight improvement over the ED features. However, the learned features based on the proposed framework gives the best performance on both the univariate and the multivariate datasets. Figure \ref{fig:scatter} shows a per dataset accuracy comparison of the proposed method with all baselines. Each point in Figure \ref{fig:scatter} represents the accuracy on one dataset from the UCR and UEA databases, with the baseline along the vertical axis and the MCL along the horizontal axis. The diagonal line indicates where two methods perform equally. Points above this line indicates that the baselines gives better performance and points below this line indicates that the MCL gives better performance. Figure \ref{fig:scatter} clearly shows that the majority of the points lie below the diagonal line, which illustrates the superior performance of the proposed method. Lastly, Figure \ref{fig:boxplot} shows a boxplot of the accuracy across all datasets in the UCR and UEA databases. The figure corroborates Table \ref{tab:results}, and illustrates that the proposed method outperforms all other baselines. 

\subsection{Transfer Learning for Clinical Time Series}

We perform transfer learning for classification of echocardiograms (ECGs) datasets with limited amount of training data, which is a typical scenario for many clinical time series datasets. First, we train an encoder using the proposed contrastive learning framework on a pretext task where a larger amount of data is available. We consider different domains for the pretext task, but with a similar amount of data. The pretext task datasets are the Syntehetic Control (Synthetic), Swedish Leaf (Dissimilar), and ECG5000 (Similar), all obtained from the UCR archive. Next, we use the weights of the encoder to initialize the weights of a supervised model, in this case the FCN, and train the model using the standard procedure. Additionally, a baseline is included where the weights are randomly initialized using He normal initialization \citep{7410480}.

\begin{figure}[htb]
    \centering
    \includegraphics[width=0.99\columnwidth]{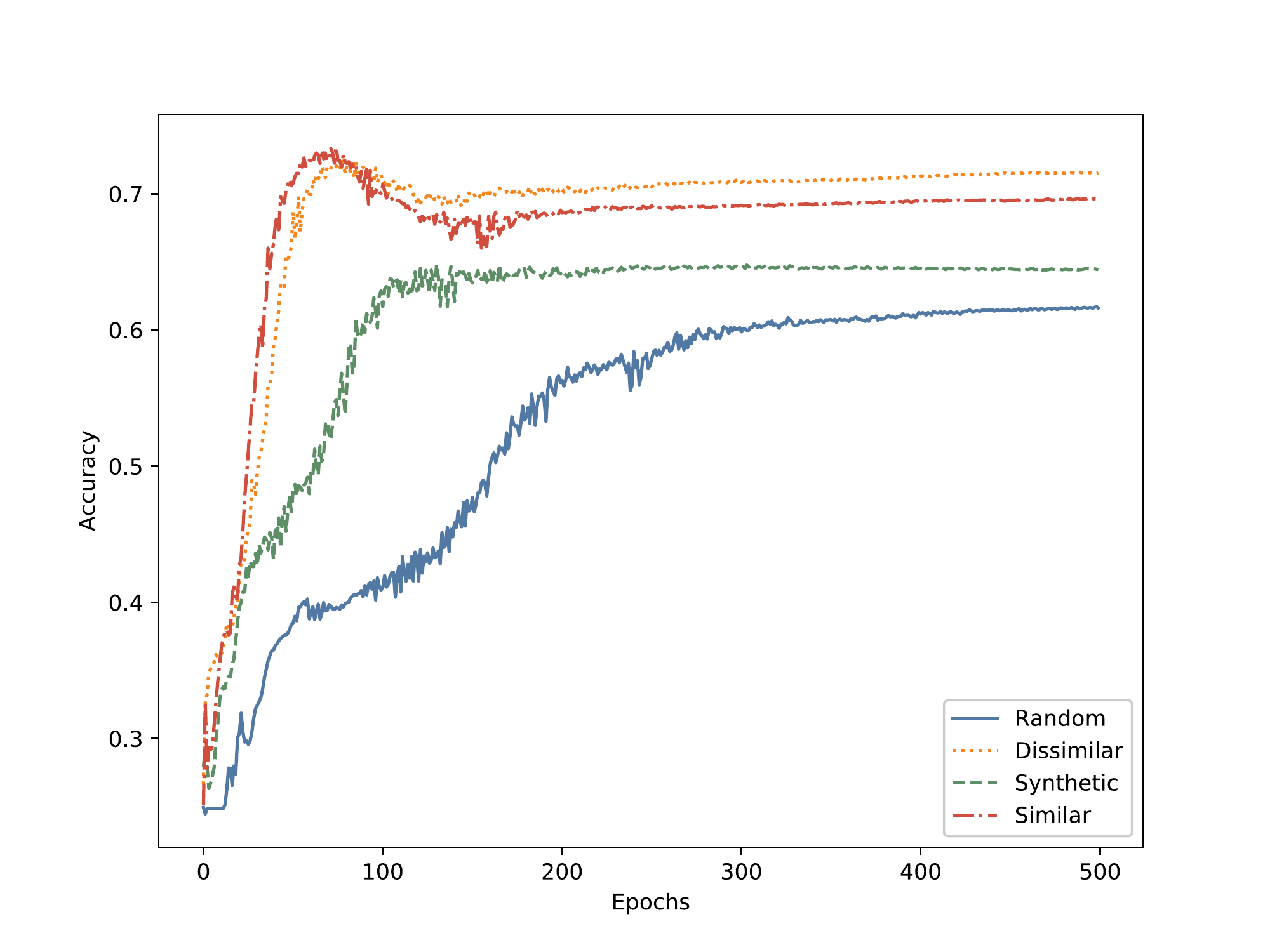}
    \caption{Accuracy of a FCN with different encoder initialization on the CinCECGTorso test data. Scores are averaged over 5 independent training runs. The figure shows that the pretrained weights using the proposed framework leads to faster convergence and increased performance.}
    \label{fig:ECGaccPlot}
\end{figure}

The results of the transfer learning experiments are presented in Table \ref{tab:resultsTL}. Using the pretrained weights obtained through the proposed contrastive learning framework leads to improved performance on most datasets. For the ECG200, the random initialization gives the highest performance. This might be a results of the ECG200 having the most training samples of the four datasets. Furthermore, Figure \ref{fig:ECGaccPlot} shows how the accuracy evolves during training, and demonstrates how using pretrained weights can lead to faster convergence and increased performance compared to random initialization. Also note that the models with weights pretrained on the similar and dissimilar domain displays a degree of overfitting after 50 epochs. At this point in the training, the loss has begun to saturate. Therefore, we believe that this overfitting might be a result of the model being to fitted to the pretext task, which hurts the performance for the down-stream task. Such challenges could be addressed through techniques such as early stopping \citep{6796297} or heavier regularization, which we consider a direction for future research.

\begin{table}[htb]
\centering
\caption{Accuracy on test data of ECG datasets with different initialization of the encoder weights. The number of training samples in each dataset is denoted by $N$. Results show how using the weights trained through the proposed contrastive framework can increase performance, particularly when the number of training samples is small.}
\vspace{0.25cm}
\resizebox{\columnwidth}{!}{%
\begin{tabular}{@{}l||c|c|c|c@{}}
\toprule
\multirow{ 2}{*}{Pretraining} & ECGFiveDays & TwoLeadECG & ECG200 & CinCECGTorso \\
 & N=23 & N=23 & N=100 & N=40 \\ \midrule \midrule
Random      &     .989$\pm$.002        & .967$\pm$.004      &    \textbf{.874}$\pm$\textbf{.008}    &   .616$\pm$.033          \\ \midrule
Synthetic   &     .997$\pm$.001       &     .985$\pm$.002       &     .866$\pm$.014       &     .644$\pm$.019       \\ \midrule
Dissimilar   &     .999$\pm$.001       &     \textbf{.987}$\pm$\textbf{.004 }      &     .868$\pm$.010       &     \textbf{.715}$\pm$\textbf{.024}       \\ \midrule
Similar     &     .998$\pm$.001       &     \textbf{.995}$\pm$\textbf{.001}       &     .842$\pm$.012       &     \textbf{.696}$\pm$\textbf{.022}       \\ \bottomrule
\end{tabular}%
}
\label{tab:resultsTL}
\end{table}

Next, the results in Table \ref{tab:resultsTL} indicate that the domain of the pretext task is important for the quality of the pretrained weights. Surprisingly, a pretext task from a dissimilar domain results in comparable results as a similar domain. It is natural to assume that a pretext task within a similar domain would be beneficial, but it is important to also consider the complexity of the data in the pretext task. In this case, the Swedish Leaf dataset is more complex as it has more classes and a more erratic nature compared to the periodic ECG5000 dataset. This might result in the encoder learning filters that can process more complicated data and generalize better to different tasks. Moreover, using the encoder trained on synthetic data also increased performance on some datasets, which indicates that useful information can be extracted even from generated data. This can be helpful for tasks with little data and no pretext task, as you can generate data and learn filters to initialize the model which might lead to a better representation.

\section{Discussion and Conclusion}

In this work, we have focused on contrastive learning of time series representations through the injection of noise, motivated by the recent success of contrastive learning on image data. However, a different line of research for contrastive learning of time series representations is using temporal information to discriminate between samples. Most recently, \citet{NIPS2019Franceschi} achieved promising results by combining a convolutional neural network encoder with a novel triplet loss, where temporal information was used to perform negative-sampling. \cite{Banville2019SelfSupervisedRL} proposed a self-supervised learning approach where an informative representation was obtained by predicting whether time windows are sampled from the same temporal context or not. \cite{NIPS2016Hyvarinen} proposed a time-contrastive learning principle that uses the non-stationary structure of the data to learn a representation where optimal discrimination of time segments is encouraged, and demonstrated how the time-contrastive learning could be related to nonlinear independent component analysis. \cite{aapo} also proposed a generalized contrastive learning framework with connections to nonlinear independent component analysis. Exploiting temporal information can be beneficial when such information is discriminative but can also encounter challenges when faced with periodic data, where noise-based approaches might succeed. We envision that our noise-based approached can be combined with temporal-based contrastive learning to reap the benefits of both approaches, and consider such a combination a promising line of future research. Lastly, a possible direction to improve the transfer learning part of our work is to include memory-based merging of features, as proposed by \citet{Ding2020}. Such an approach could allow for samples from the source and target domain to be merged and potentially increase performance.

This paper introduced a novel self-supervised framework for time series representation learning. The framework exploits a recent augmentation technique called miuxp, in which new samples are generated through combinations of data points. The proposed framework was evaluated on numerous datasets with encouraging results. Furthermore, we demonstrated how the proposed framework enables transfer learning for clinical time series with good results. We believe that our proposed framework can be a useful approach for time series representation learning.

\section{Acknowledgement}

This work was financially supported by the Research Council of Norway (RCN), through its Centre for Research-based Innovation funding scheme (Visual Intelligence, grant no. 309439), and Consortium Partners. The work was further partially funded by RCN FRIPRO grant no. 315029, RCN IKTPLUSS grant no. 303514, and the UiT Thematic Initiative “Data-Driven Health Technology”.

\appendix\label{appendix}

\section{Results on individual datasets}

Table \ref{tab:UCRallResults1}, \ref{tab:UCRallResults2}, and \ref{tab:UEAallResults} displays the accuracy of all methods evaluated in the article on all datasets of the UCR and UEA databases, respectively. For the learning-based methods (AE, CL, and MCL), the scores represent the average accuracy across five independent training runs. Results for all 5 training runs and ranks on individual datasets are available at \url{https://github.com/Wickstrom/MixupContrastiveLearning} along with code.

\begin{table}[htb]
\centering
\caption{Accuracy of a 1NN classifier on UCR datasets starting with the letters A-R. For AE, CLG, CLD, and MCL, the results are the average across 5 different training runs. Datasets from A-R.}
\vspace{0.25cm}
\resizebox{0.75\columnwidth}{!}{%
\begin{tabular}{lcccccc}
\toprule
Dataset               & HC & ED & AE & $CL(\rho=.25)$ & $CL(\sigma=.25)$ & MCL \\ \midrule
ACSF1                          &  0.58 &  0.54 &  0.50 &              0.76 &            0.75 &  0.90 \\
Adiac                          &  0.41 &  0.61 &  0.58 &              0.54 &            0.55 &  0.68 \\
AllGestureWiimoteX             &  0.37 &  0.52 &  0.58 &              0.53 &            0.51 &  0.66 \\
AllGestureWiimoteY             &  0.38 &  0.57 &  0.55 &              0.58 &            0.55 &  0.73 \\
AllGestureWiimoteZ             &  0.30 &  0.45 &  0.46 &              0.41 &            0.41 &  0.62 \\
ArrowHead                      &  0.65 &  0.80 &  0.81 &              0.61 &            0.63 &  0.82 \\
BME                            &  0.56 &  0.83 &  0.83 &              0.67 &            0.63 &  0.98 \\
Beef                           &  0.50 &  0.67 &  0.67 &              0.42 &            0.43 &  0.67 \\
BeetleFly                      &  0.95 &  0.75 &  0.75 &              0.76 &            0.72 &  0.75 \\
BirdChicken                    &  0.70 &  0.55 &  0.73 &              0.94 &            0.94 &  0.82 \\
CBF                            &  0.66 &  0.85 &  0.95 &              0.99 &            1.00 &  0.94 \\
Car                            &  0.42 &  0.73 &  0.73 &              0.48 &            0.48 &  0.78 \\
Chinatown                      &  0.51 &  0.95 &  0.88 &              0.71 &            0.87 &  0.93 \\
ChlorineConcentration          &  0.45 &  0.65 &  0.53 &              0.46 &            0.45 &  0.66 \\
CinCECGTorso                   &  0.50 &  0.90 &  0.90 &              0.52 &            0.52 &  0.72 \\
Coffee                         &  0.57 &  1.00 &  1.00 &              0.96 &            0.96 &  0.94 \\
Computers                      &  0.58 &  0.58 &  0.58 &              0.60 &            0.61 &  0.67 \\
CricketX                       &  0.26 &  0.58 &  0.56 &              0.55 &            0.53 &  0.71 \\
CricketY                       &  0.18 &  0.57 &  0.55 &              0.54 &            0.47 &  0.68 \\
CricketZ                       &  0.26 &  0.59 &  0.56 &              0.60 &            0.52 &  0.72 \\
Crop                           &  0.48 &  0.71 &  0.70 &              0.54 &            0.56 &  0.73 \\
DiatomSizeReduction            &  0.99 &  0.93 &  0.94 &              0.86 &            0.84 &  0.87 \\
DistalPhalanxOutlineAgeGroup   &  0.61 &  0.63 &  0.64 &              0.69 &            0.68 &  0.62 \\
DistalPhalanxOutlineCorrect    &  0.67 &  0.72 &  0.73 &              0.71 &            0.71 &  0.66 \\
DistalPhalanxTW                &  0.49 &  0.63 &  0.60 &              0.60 &            0.58 &  0.55 \\
DodgerLoopDay                  &  0.31 &  0.55 &  0.57 &              0.41 &            0.44 &  0.49 \\
DodgerLoopGame                 &  0.59 &  0.88 &  0.85 &              0.72 &            0.69 &  0.79 \\
DodgerLoopWeekend              &  0.81 &  0.99 &  0.99 &              0.91 &            0.87 &  0.95 \\
ECG200                         &  0.72 &  0.88 &  0.90 &              0.80 &            0.77 &  0.87 \\
ECG5000                        &  0.85 &  0.92 &  0.93 &              0.92 &            0.92 &  0.92 \\
ECGFiveDays                    &  0.73 &  0.80 &  0.82 &              0.86 &            0.91 &  0.94 \\
EOGHorizontalSignal            &  0.29 &  0.42 &  0.45 &              0.36 &            0.36 &  0.44 \\
EOGVerticalSignal              &  0.17 &  0.44 &  0.38 &              0.26 &            0.25 &  0.38 \\
Earthquakes                    &  0.64 &  0.71 &  0.70 &              0.63 &            0.64 &  0.69 \\
ElectricDevices                &  0.41 &  0.55 &  0.56 &              0.54 &            0.55 &  0.61 \\
EthanolLevel                   &  0.27 &  0.27 &  0.29 &              0.32 &            0.32 &  0.51 \\
FaceAll                        &  0.21 &  0.71 &  0.69 &              0.66 &            0.67 &  0.79 \\
FaceFour                       &  0.41 &  0.78 &  0.79 &              0.63 &            0.81 &  0.85 \\
FacesUCR                       &  0.34 &  0.77 &  0.77 &              0.76 &            0.79 &  0.91 \\
FiftyWords                     &  0.13 &  0.63 &  0.59 &              0.38 &            0.37 &  0.60 \\
Fish                           &  0.27 &  0.78 &  0.80 &              0.65 &            0.64 &  0.85 \\
FordA                          &  0.53 &  0.67 &  0.67 &              0.81 &            0.79 &  0.88 \\
FordB                          &  0.51 &  0.61 &  0.61 &              0.68 &            0.68 &  0.73 \\
FreezerRegularTrain            &  0.94 &  0.80 &  0.88 &              0.90 &            0.90 &  0.96 \\
FreezerSmallTrain              &  0.88 &  0.68 &  0.70 &              0.69 &            0.69 &  0.79 \\
Fungi                          &  0.39 &  0.82 &  0.82 &              0.70 &            0.69 &  0.93 \\
GestureMidAirD1                &  0.13 &  0.58 &  0.58 &              0.29 &            0.28 &  0.56 \\
GestureMidAirD2                &  0.10 &  0.49 &  0.45 &              0.30 &            0.28 &  0.51 \\
GestureMidAirD3                &  0.11 &  0.35 &  0.34 &              0.18 &            0.16 &  0.30 \\
GesturePebbleZ1                &  0.34 &  0.73 &  0.71 &              0.69 &            0.68 &  0.68 \\
GesturePebbleZ2                &  0.35 &  0.67 &  0.63 &              0.58 &            0.58 &  0.65 \\
GunPoint                       &  0.74 &  0.91 &  0.93 &              0.85 &            0.85 &  1.00 \\
GunPointAgeSpan                &  0.71 &  0.90 &  0.98 &              0.95 &            0.96 &  0.99 \\
GunPointMaleVersusFemale       &  0.82 &  0.97 &  1.00 &              0.98 &            0.96 &  1.00 \\
GunPointOldVersusYoung         &  1.00 &  0.95 &  1.00 &              1.00 &            1.00 &  1.00 \\
Ham                            &  0.45 &  0.60 &  0.52 &              0.52 &            0.56 &  0.57 \\
HandOutlines                   &  0.63 &  0.86 &  0.86 &              0.70 &            0.69 &  0.80 \\
Haptics                        &  0.26 &  0.37 &  0.36 &              0.31 &            0.30 &  0.43 \\
Herring                        &  0.55 &  0.52 &  0.56 &              0.51 &            0.50 &  0.57 \\
HouseTwenty                    &  0.47 &  0.66 &  0.64 &              0.87 &            0.77 &  0.89 \\
InlineSkate                    &  0.27 &  0.34 &  0.32 &              0.29 &            0.30 &  0.41 \\
InsectEPGRegularTrain          &  1.00 &  0.68 &  1.00 &              1.00 &            1.00 &  1.00 \\
InsectEPGSmallTrain            &  1.00 &  0.66 &  1.00 &              1.00 &            1.00 &  1.00 \\
InsectWingbeatSound            &  0.13 &  0.56 &  0.54 &              0.35 &            0.32 &  0.42 \\
ItalyPowerDemand               &  0.61 &  0.96 &  0.94 &              0.91 &            0.93 &  0.94 \\
LargeKitchenAppliances         &  0.56 &  0.49 &  0.47 &              0.73 &            0.74 &  0.74 \\
Lightning2                     &  0.56 &  0.75 &  0.77 &              0.76 &            0.75 &  0.78 \\
Lightning7                     &  0.42 &  0.58 &  0.62 &              0.55 &            0.58 &  0.72 \\
Mallat                         &  0.38 &  0.91 &  0.92 &              0.85 &            0.84 &  0.83 \\
Meat                           &  0.48 &  0.93 &  0.93 &              0.87 &            0.86 &  0.84 \\
MedicalImages                  &  0.43 &  0.68 &  0.64 &              0.62 &            0.61 &  0.68 \\
MelbournePedestrian            &  0.42 &  0.85 &  0.94 &              0.85 &            0.85 &  0.93 \\
MiddlePhalanxOutlineAgeGroup   &  0.47 &  0.52 &  0.54 &              0.49 &            0.49 &  0.48 \\
MiddlePhalanxOutlineCorrect    &  0.64 &  0.77 &  0.75 &              0.73 &            0.73 &  0.67 \\
MiddlePhalanxTW                &  0.42 &  0.51 &  0.49 &              0.45 &            0.47 &  0.49 \\
MixedShapesRegularTrain        &  0.43 &  0.90 &  0.90 &              0.60 &            0.60 &  0.92 \\
MixedShapesSmallTrain          &  0.44 &  0.84 &  0.83 &              0.54 &            0.53 &  0.84 \\
MoteStrain                     &  0.74 &  0.88 &  0.84 &              0.85 &            0.84 &  0.88 \\
NonInvasiveFetalECGThorax1     &  0.34 &  0.83 &  0.82 &              0.57 &            0.61 &  0.84 \\
NonInvasiveFetalECGThorax2     &  0.40 &  0.88 &  0.87 &              0.68 &            0.69 &  0.88 \\
OSULeaf                        &  0.37 &  0.52 &  0.51 &              0.66 &            0.61 &  0.87 \\
OliveOil                       &  0.33 &  0.87 &  0.87 &              0.71 &            0.70 &  0.74 \\
PLAID                          &  0.70 &  0.52 &  0.72 &              0.56 &            0.54 &  0.76 \\
PhalangesOutlinesCorrect       &  0.64 &  0.76 &  0.74 &              0.71 &            0.73 &  0.71 \\
Phoneme                        &  0.09 &  0.11 &  0.12 &              0.19 &            0.19 &  0.22 \\
PickupGestureWiimoteZ          &  0.48 &  0.56 &  0.74 &              0.55 &            0.56 &  0.67 \\
PigAirwayPressure              &  0.51 &  0.06 &  0.09 &              0.33 &            0.30 &  0.31 \\
PigArtPressure                 &  0.69 &  0.12 &  0.27 &              0.72 &            0.62 &  0.96 \\
PigCVP                         &  0.63 &  0.08 &  0.14 &              0.44 &            0.42 &  0.85 \\
Plane                          &  0.80 &  0.96 &  0.97 &              0.97 &            0.96 &  0.99 \\
PowerCons                      &  0.92 &  0.93 &  0.98 &              0.94 &            0.93 &  0.90 \\
ProximalPhalanxOutlineAgeGroup &  0.72 &  0.79 &  0.79 &              0.78 &            0.78 &  0.76 \\
ProximalPhalanxOutlineCorrect  &  0.68 &  0.81 &  0.78 &              0.75 &            0.77 &  0.80 \\
ProximalPhalanxTW              &  0.60 &  0.71 &  0.71 &              0.70 &            0.68 &  0.66 \\
RefrigerationDevices           &  0.45 &  0.39 &  0.39 &              0.49 &            0.46 &  0.48 \\
Rock                           &  0.42 &  0.84 &  0.72 &              0.42 &            0.41 &  0.60 \\\bottomrule
\end{tabular}}
\label{tab:UCRallResults1}
\end{table}
\begin{table}[htb]
\centering
\caption{Accuracy of a 1NN classifier on UCR datasets starting with the letters S-Y. For AE, CLG, CLD, and MCL, the results are the average across 5 different training runs.}
\vspace{0.25cm}
\resizebox{0.75\columnwidth}{!}{%
\begin{tabular}{lcccccc}
\toprule
Dataset               & HC & ED & AE & $CL(\rho=.25)$ & $CL(\sigma=.25)$ & MCL \\ \midrule
ScreenType                     &  0.39 &  0.36 &  0.37 &              0.41 &            0.42 &  0.48 \\
SemgHandGenderCh2              &  0.78 &  0.76 &  0.91 &              0.79 &            0.77 &  0.83 \\
SemgHandMovementCh2            &  0.47 &  0.37 &  0.69 &              0.58 &            0.56 &  0.60 \\
SemgHandSubjectCh2             &  0.56 &  0.40 &  0.84 &              0.68 &            0.66 &  0.68 \\
ShakeGestureWiimoteZ           &  0.62 &  0.60 &  0.81 &              0.84 &            0.83 &  0.91 \\
ShapeletSim                    &  0.45 &  0.54 &  0.54 &              0.79 &            0.75 &  0.83 \\
ShapesAll                      &  0.30 &  0.75 &  0.73 &              0.64 &            0.63 &  0.84 \\
SmallKitchenAppliances         &  0.52 &  0.34 &  0.39 &              0.67 &            0.67 &  0.71 \\
SmoothSubspace                 &  0.81 &  0.91 &  0.81 &              0.87 &            0.87 &  0.92 \\
SonyAIBORobotSurface1          &  0.64 &  0.70 &  0.67 &              0.79 &            0.74 &  0.67 \\
SonyAIBORobotSurface2          &  0.65 &  0.86 &  0.85 &              0.83 &            0.85 &  0.83 \\
StarLightCurves                &  0.85 &  0.85 &  0.86 &              0.86 &            0.85 &  0.97 \\
Strawberry                     &  0.70 &  0.95 &  0.94 &              0.86 &            0.86 &  0.96 \\
SwedishLeaf                    &  0.34 &  0.79 &  0.79 &              0.86 &            0.84 &  0.90 \\
Symbols                        &  0.39 &  0.90 &  0.89 &              0.83 &            0.77 &  0.94 \\
SyntheticControl               &  0.42 &  0.88 &  0.93 &              0.98 &            0.99 &  0.95 \\
ToeSegmentation1               &  0.63 &  0.68 &  0.69 &              0.80 &            0.78 &  0.90 \\
ToeSegmentation2               &  0.71 &  0.81 &  0.79 &              0.85 &            0.79 &  0.90 \\
Trace                          &  1.00 &  0.76 &  0.80 &              0.89 &            0.86 &  1.00 \\
TwoLeadECG                     &  0.67 &  0.75 &  0.70 &              0.75 &            0.72 &  0.90 \\
TwoPatterns                    &  0.28 &  0.91 &  0.92 &              0.97 &            0.96 &  0.88 \\
UMD                            &  0.94 &  0.76 &  0.76 &              0.86 &            0.85 &  0.97 \\
UWaveGestureLibraryAll         &  0.19 &  0.95 &  0.94 &              0.46 &            0.44 &  0.76 \\
UWaveGestureLibraryX           &  0.22 &  0.74 &  0.73 &              0.54 &            0.49 &  0.74 \\
UWaveGestureLibraryY           &  0.21 &  0.66 &  0.63 &              0.48 &            0.44 &  0.67 \\
UWaveGestureLibraryZ           &  0.22 &  0.65 &  0.64 &              0.53 &            0.50 &  0.70 \\
Wafer                          &  0.95 &  1.00 &  0.99 &              0.98 &            0.98 &  0.99 \\
Wine                           &  0.48 &  0.61 &  0.63 &              0.64 &            0.61 &  0.61 \\
WordSynonyms                   &  0.17 &  0.62 &  0.58 &              0.39 &            0.39 &  0.61 \\
Worms                          &  0.56 &  0.45 &  0.43 &              0.60 &            0.56 &  0.78 \\
WormsTwoClass                  &  0.65 &  0.61 &  0.62 &              0.64 &            0.64 &  0.82 \\
Yoga                           &  0.60 &  0.83 &  0.80 &              0.75 &            0.76 &  0.79 \\ \bottomrule
\end{tabular}}

\label{tab:UCRallResults2}
\end{table}
\begin{table}[htb]
\centering
\caption{Accuracy of a 1NN classifier on all UEA datasets. For AE, CLG, CLD, and MCL, the results are the average across 5 different training runs.}
\vspace{0.25cm}
\resizebox{0.75\columnwidth}{!}{%
\begin{tabular}{lcccccc}
\toprule
Dataset               & HC & ED & AE & $CL(\rho=.25)$ & $CL(\sigma=.25)$ & MCL \\ \midrule
ArticularyWordRecognition &  0.78 &  0.97 &  0.97 &              0.87 &            0.90 &  0.97 \\
AtrialFibrillation        &  0.13 &  0.27 &  0.24 &              0.23 &            0.31 &  0.17 \\
BasicMotions              &  1.00 &  0.68 &  0.97 &              1.00 &            0.98 &  1.00 \\
CharacterTrajectories     &  0.82 &  0.96 &  0.94 &              0.94 &            0.90 &  0.98 \\
Cricket                   &  0.92 &  0.94 &  0.90 &              0.91 &            0.94 &  0.96 \\
DuckDuckGeese             &  0.50 &  0.28 &  0.41 &              0.36 &            0.34 &  0.47 \\
ERing                     &  0.67 &  0.13 &  0.91 &              0.83 &            0.83 &  0.87 \\
EigenWorms                &  0.66 &  0.55 &  0.00 &              0.61 &            0.62 &  0.71 \\
Epilepsy                  &  0.97 &  0.67 &  0.83 &              0.93 &            0.93 &  0.96 \\
EthanolConcentration      &  0.27 &  0.29 &  0.28 &              0.30 &            0.29 &  0.28 \\
FaceDetection             &  0.51 &  0.52 &  0.52 &              0.50 &            0.50 &  0.50 \\
FingerMovements           &  0.52 &  0.55 &  0.52 &              0.53 &            0.51 &  0.61 \\
HandMovementDirection     &  0.26 &  0.28 &  0.28 &              0.25 &            0.29 &  0.35 \\
Handwriting               &  0.11 &  0.20 &  0.34 &              0.43 &            0.42 &  0.52 \\
Heartbeat                 &  0.65 &  0.62 &  0.70 &              0.69 &            0.70 &  0.68 \\
InsectWingbeat            &  0.00 &  0.00 &  0.00 &              0.00 &            0.00 &  0.00 \\
JapaneseVowels            &  0.96 &  0.92 &  0.92 &              0.87 &            0.88 &  0.87 \\
LSST                      &  0.55 &  0.46 &  0.45 &              0.43 &            0.39 &  0.44 \\
Libras                    &  0.61 &  0.83 &  0.78 &              0.61 &            0.57 &  0.89 \\
MotorImagery              &  0.46 &  0.51 &  0.53 &              0.56 &            0.55 &  0.56 \\
NATOPS                    &  0.66 &  0.85 &  0.84 &              0.76 &            0.73 &  0.82 \\
PEMS-SF                   &  0.66 &  0.70 &  0.79 &              0.80 &            0.80 &  0.71 \\
PenDigits                 &  0.53 &  0.97 &  0.00 &              0.86 &            0.87 &  0.97 \\
Phoneme                   &  0.07 &  0.10 &  0.12 &              0.07 &            0.06 &  0.16 \\
RacketSports              &  0.75 &  0.87 &  0.79 &              0.80 &            0.79 &  0.82 \\
SelfRegulationSCP1        &  0.77 &  0.77 &  0.78 &              0.72 &            0.72 &  0.68 \\
SelfRegulationSCP2        &  0.52 &  0.48 &  0.49 &              0.52 &            0.50 &  0.52 \\
SpokenArabicDigits        &  0.76 &  0.97 &  0.94 &              0.46 &            0.50 &  0.93 \\
StandWalkJump             &  0.33 &  0.20 &  0.56 &              0.31 &            0.36 &  0.27 \\
UWaveGestureLibrary       &  0.37 &  0.88 &  0.83 &              0.61 &            0.60 &  0.87 \\ \bottomrule
\end{tabular}}

\label{tab:UEAallResults}
\end{table}






\bibliographystyle{elsarticle-num-names}
\bibliography{bibliography}

\begin{thebibliography}{28}
\expandafter\ifx\csname natexlab\endcsname\relax\def\natexlab#1{#1}\fi
\providecommand{\url}[1]{\texttt{#1}}
\providecommand{\href}[2]{#2}
\providecommand{\path}[1]{#1}
\providecommand{\DOIprefix}{doi:}
\providecommand{\ArXivprefix}{arXiv:}
\providecommand{\URLprefix}{URL: }
\providecommand{\Pubmedprefix}{pmid:}
\providecommand{\doi}[1]{\href{http://dx.doi.org/#1}{\path{#1}}}
\providecommand{\Pubmed}[1]{\href{pmid:#1}{\path{#1}}}
\providecommand{\bibinfo}[2]{#2}
\ifx\xfnm\relax \def\xfnm[#1]{\unskip,\space#1}\fi
\bibitem[{Ching et~al.(2018)Ching, Himmelstein, Beaulieu-Jones
  et~al.}]{missdata}
\bibinfo{author}{T.~Ching}, \bibinfo{author}{D.~S. Himmelstein},
  \bibinfo{author}{B.~K. Beaulieu-Jones}, et~al.,
\newblock \bibinfo{title}{Opportunities and obstacles for deep learning in
  biology and medicine},
\newblock \bibinfo{journal}{Journal of The Royal Society Interface}
  (\bibinfo{year}{2018}) \bibinfo{pages}{20170387}.
  \DOIprefix\doi{10.1098/rsif.2017.0387}.
\bibitem[{{Pan} and {Yang}(2010)}]{5288526}
\bibinfo{author}{S.~J. {Pan}}, \bibinfo{author}{Q.~{Yang}},
\newblock \bibinfo{title}{A survey on transfer learning},
\newblock \bibinfo{journal}{IEEE Transactions on Knowledge and Data
  Engineering}  (\bibinfo{year}{2010}) \bibinfo{pages}{1345--1359}.
\bibitem[{Bengio et~al.(2013)Bengio, Courville, and
  Vincent}]{Bengio2013RepresentationLA}
\bibinfo{author}{Y.~Bengio}, \bibinfo{author}{A.~C. Courville},
  \bibinfo{author}{P.~Vincent},
\newblock \bibinfo{title}{Representation learning: A review and new
  perspectives},
\newblock \bibinfo{journal}{IEEE Transactions on Pattern Analysis and Machine
  Intelligence} \bibinfo{volume}{35} (\bibinfo{year}{2013})
  \bibinfo{pages}{1798--1828}.
\bibitem[{Jing and Tian(2019)}]{9086055}
\bibinfo{author}{L.~Jing}, \bibinfo{author}{Y.~Tian},
\newblock \bibinfo{title}{Self-supervised visual feature learning with deep
  neural networks: A survey},
\newblock \bibinfo{journal}{IEEE Transactions on Pattern Analysis and Machine
  Intelligence}  (\bibinfo{year}{2019}) \bibinfo{pages}{1--1}.
  \DOIprefix\doi{10.1109/TPAMI.2020.2992393}.
\bibitem[{Chen et~al.(2020)Chen, Kornblith, Norouzi, and Hinton}]{Chen2020ASF}
\bibinfo{author}{T.~Chen}, \bibinfo{author}{S.~Kornblith},
  \bibinfo{author}{M.~Norouzi}, \bibinfo{author}{G.~Hinton},
\newblock \bibinfo{title}{A simple framework for contrastive learning of visual
  representations},
\newblock in: \bibinfo{booktitle}{International Conference on Machine
  Learning}, \bibinfo{year}{2020}, pp. \bibinfo{pages}{1597--1607}.
\bibitem[{He et~al.(2020)He, Fan, Wu, Xie, and Girshick}]{MoCocvpr}
\bibinfo{author}{K.~He}, \bibinfo{author}{H.~Fan}, \bibinfo{author}{Y.~Wu},
  \bibinfo{author}{S.~Xie}, \bibinfo{author}{R.~Girshick},
\newblock \bibinfo{title}{Momentum contrast for unsupervised visual
  representation learning},
\newblock in: \bibinfo{booktitle}{Conference on Computer Vision and Pattern
  Recognition}, \bibinfo{year}{2020}, pp. \bibinfo{pages}{9729--9738}.
\bibitem[{Grill et~al.(2020)Grill, Strub, Altch\'{e} et~al.}]{ByolGrill}
\bibinfo{author}{J.-B. Grill}, \bibinfo{author}{F.~Strub},
  \bibinfo{author}{F.~Altch\'{e}}, et~al.,
\newblock \bibinfo{title}{Bootstrap your own latent: A new approach to
  self-supervised learning},
\newblock in: \bibinfo{booktitle}{Advances in Neural Information Processing
  Systems}, \bibinfo{year}{2020}, pp. \bibinfo{pages}{21271--21284}.
\bibitem[{Shorten and Khoshgoftaar(2019)}]{Shorten2019ASO}
\bibinfo{author}{C.~Shorten}, \bibinfo{author}{T.~Khoshgoftaar},
\newblock \bibinfo{title}{A survey on image data augmentation for deep
  learning},
\newblock \bibinfo{journal}{Journal of Big Data} \bibinfo{volume}{6}
  (\bibinfo{year}{2019}) \bibinfo{pages}{1--48}.
\bibitem[{Zhang et~al.(2018)Zhang, Ciss{\'{e}}, Dauphin, and
  Lopez{-}Paz}]{zhang2018mixup}
\bibinfo{author}{H.~Zhang}, \bibinfo{author}{M.~Ciss{\'{e}}},
  \bibinfo{author}{Y.~N. Dauphin}, \bibinfo{author}{D.~Lopez{-}Paz},
\newblock \bibinfo{title}{mixup: Beyond empirical risk minimization},
\newblock in: \bibinfo{booktitle}{International Conference on Learning
  Representations}, \bibinfo{year}{2018}.
\bibitem[{M\"{u}ller et~al.(2019)M\"{u}ller, Kornblith, and Hinton}]{labelS}
\bibinfo{author}{R.~M\"{u}ller}, \bibinfo{author}{S.~Kornblith},
  \bibinfo{author}{G.~E. Hinton},
\newblock \bibinfo{title}{When does label smoothing help?},
\newblock in: \bibinfo{booktitle}{Advances in Neural Information Processing
  Systems}, \bibinfo{year}{2019}, pp. \bibinfo{pages}{4694--4703}.
\bibitem[{Dau et~al.(2018)Dau, Bagnall, Kamgar et~al.}]{UTS}
\bibinfo{author}{H.~A. Dau}, \bibinfo{author}{A.~J. Bagnall},
  \bibinfo{author}{K.~Kamgar}, et~al., \bibinfo{title}{The {UCR} time series
  archive}, \bibinfo{year}{2018}. \URLprefix
  \url{http://arxiv.org/abs/1810.07758}.
  \href{http://arxiv.org/abs/1810.07758}{{\tt arXiv:1810.07758}}.
\bibitem[{Bagnall et~al.(2018)Bagnall, Dau, Lines et~al.}]{MTS}
\bibinfo{author}{A.~J. Bagnall}, \bibinfo{author}{H.~A. Dau},
  \bibinfo{author}{J.~Lines}, et~al.,
\newblock \bibinfo{title}{The {UEA} multivariate time series classification
  archive, 2018},
\newblock \bibinfo{journal}{CoRR} \bibinfo{volume}{abs/1811.00075}
  (\bibinfo{year}{2018}).
\bibitem[{Oh et~al.(2018)Oh, Wang, and Wiens}]{pmlrJeeheh}
\bibinfo{author}{J.~Oh}, \bibinfo{author}{J.~Wang}, \bibinfo{author}{J.~Wiens},
\newblock \bibinfo{title}{Learning to exploit invariances in clinical
  time-series data using sequence transformer networks},
\newblock in: \bibinfo{booktitle}{Proceedings of Machine Learning Research},
  volume~\bibinfo{volume}{85}, \bibinfo{address}{Palo Alto, California},
  \bibinfo{year}{2018}, pp. \bibinfo{pages}{332--347}.
\bibitem[{Hein et~al.(2019)Hein, Andriushchenko, and
  Bitterwolf}]{Hein2019WhyRN}
\bibinfo{author}{M.~Hein}, \bibinfo{author}{M.~Andriushchenko},
  \bibinfo{author}{J.~Bitterwolf},
\newblock \bibinfo{title}{Why relu networks yield high-confidence predictions
  far away from the training data and how to mitigate the problem},
\newblock \bibinfo{journal}{Conference on Computer Vision and Pattern
  Recognition}  (\bibinfo{year}{2019}) \bibinfo{pages}{41--50}.
\bibitem[{{Zhang} et~al.(2017){Zhang}, {Isola}, and {Efros}}]{8099559}
\bibinfo{author}{R.~{Zhang}}, \bibinfo{author}{P.~{Isola}},
  \bibinfo{author}{A.~A. {Efros}},
\newblock \bibinfo{title}{Split-brain autoencoders: Unsupervised learning by
  cross-channel prediction},
\newblock in: \bibinfo{booktitle}{Conference on Computer Vision and Pattern
  Recognition}, \bibinfo{year}{2017}, pp. \bibinfo{pages}{645--654}.
\bibitem[{Caron et~al.(2018)Caron, Bojanowski, Joulin et~al.}]{Caron}
\bibinfo{author}{M.~Caron}, \bibinfo{author}{P.~Bojanowski},
  \bibinfo{author}{A.~Joulin}, et~al.,
\newblock \bibinfo{title}{Deep clustering for unsupervised learning of visual
  features},
\newblock in: \bibinfo{booktitle}{Proceedings of the European Conference on
  Computer Vision}, \bibinfo{year}{2018}, pp. \bibinfo{pages}{139--156}.
\bibitem[{Zhu et~al.(2021)Zhu, Fan, Luo, Xu, and Yang}]{9349150}
\bibinfo{author}{L.~Zhu}, \bibinfo{author}{H.~Fan}, \bibinfo{author}{Y.~Luo},
  \bibinfo{author}{M.~Xu}, \bibinfo{author}{Y.~Yang},
\newblock \bibinfo{title}{Temporal cross-layer correlation mining for action
  recognition},
\newblock \bibinfo{journal}{IEEE Transactions on Multimedia}
  (\bibinfo{year}{2021}) \bibinfo{pages}{1--1}.
  \DOIprefix\doi{10.1109/TMM.2021.3057503}.
\bibitem[{Kieu et~al.(2019)Kieu, Yang, Guo, and Jensen}]{ijcaiex2}
\bibinfo{author}{T.~Kieu}, \bibinfo{author}{B.~Yang}, \bibinfo{author}{C.~Guo},
  \bibinfo{author}{C.~S. Jensen},
\newblock \bibinfo{title}{Outlier detection for time series with recurrent
  autoencoder ensembles},
\newblock in: \bibinfo{booktitle}{International Joint Conference on Artificial
  Intelligence}, \bibinfo{year}{2019}, pp. \bibinfo{pages}{2725--2732}.
\bibitem[{{Wang} et~al.(2017){Wang}, {Yan}, and {Oates}}]{7966039}
\bibinfo{author}{Z.~{Wang}}, \bibinfo{author}{W.~{Yan}},
  \bibinfo{author}{T.~{Oates}},
\newblock \bibinfo{title}{Time series classification from scratch with deep
  neural networks: A strong baseline},
\newblock in: \bibinfo{booktitle}{Proceedings of the International Joint
  Conference on Neural Networks}, \bibinfo{year}{2017}, pp.
  \bibinfo{pages}{1578--1585}.
\bibitem[{Ioffe and Szegedy(2015)}]{batchnorm}
\bibinfo{author}{S.~Ioffe}, \bibinfo{author}{C.~Szegedy},
\newblock \bibinfo{title}{Batch normalization: Accelerating deep network
  training by reducing internal covariate shift},
\newblock in: \bibinfo{booktitle}{International Conference on Machine
  Learning}, \bibinfo{year}{2015}, p. \bibinfo{pages}{448–456}.
\bibitem[{Kingma and Ba(2014)}]{ADAM}
\bibinfo{author}{D.~Kingma}, \bibinfo{author}{J.~Ba},
\newblock \bibinfo{title}{Adam: A method for stochastic optimization},
\newblock in: \bibinfo{booktitle}{International Conference on Learning
  Representations}, \bibinfo{year}{2014}.
\bibitem[{{He} et~al.(2015){He}, {Zhang}, Ren et~al.}]{7410480}
\bibinfo{author}{K.~{He}}, \bibinfo{author}{X.~{Zhang}},
  \bibinfo{author}{S.~Ren}, et~al.,
\newblock \bibinfo{title}{Delving deep into rectifiers: Surpassing human-level
  performance on imagenet classification},
\newblock in: \bibinfo{booktitle}{2015 IEEE International Conference on
  Computer Vision}, \bibinfo{year}{2015}, pp. \bibinfo{pages}{1026--1034}.
\bibitem[{{Girosi} et~al.(1995){Girosi}, {Jones}, and {Poggio}}]{6796297}
\bibinfo{author}{F.~{Girosi}}, \bibinfo{author}{M.~{Jones}},
  \bibinfo{author}{T.~{Poggio}},
\newblock \bibinfo{title}{Regularization theory and neural networks
  architectures},
\newblock \bibinfo{journal}{Neural Computation}  (\bibinfo{year}{1995})
  \bibinfo{pages}{219--269}.
\bibitem[{Franceschi et~al.(2019)Franceschi, Dieuleveut, and
  Jaggi}]{NIPS2019Franceschi}
\bibinfo{author}{J.-Y. Franceschi}, \bibinfo{author}{A.~Dieuleveut},
  \bibinfo{author}{M.~Jaggi},
\newblock \bibinfo{title}{Unsupervised scalable representation learning for
  multivariate time series},
\newblock in: \bibinfo{booktitle}{Advances in Neural Information Processing
  Systems}, \bibinfo{year}{2019}, pp. \bibinfo{pages}{4650--4661}.
\bibitem[{Banville et~al.(2019)Banville, Albuquerque, and
  Hyvarinen}]{Banville2019SelfSupervisedRL}
\bibinfo{author}{H.~J. Banville}, \bibinfo{author}{I.~Albuquerque},
  \bibinfo{author}{A.~Hyvarinen},
\newblock \bibinfo{title}{Self-supervised representation learning from
  electroencephalography signals},
\newblock \bibinfo{journal}{International Workshop on Machine Learning for
  Signal Processing}  (\bibinfo{year}{2019}) \bibinfo{pages}{1--6}.
\bibitem[{Hyvarinen and Morioka(2016)}]{NIPS2016Hyvarinen}
\bibinfo{author}{A.~Hyvarinen}, \bibinfo{author}{H.~Morioka},
\newblock \bibinfo{title}{Unsupervised feature extraction by time-contrastive
  learning and nonlinear ica},
\newblock in: \bibinfo{booktitle}{Advances in Neural Information Processing
  Systems}, \bibinfo{year}{2016}, pp. \bibinfo{pages}{3765--3773}.
\bibitem[{Hyv{\"a}rinen et~al.(2019)Hyv{\"a}rinen, Sasaki, and Turner}]{aapo}
\bibinfo{author}{A.~Hyv{\"a}rinen}, \bibinfo{author}{H.~Sasaki},
  \bibinfo{author}{R.~Turner},
\newblock \bibinfo{title}{Nonlinear ica using auxiliary variables and
  generalized contrastive learning},
\newblock in: \bibinfo{booktitle}{The 22nd International Conference on
  Artificial Intelligence and Statistics}, \bibinfo{year}{2019}, pp.
  \bibinfo{pages}{859--868}.
\bibitem[{Ding et~al.(2020)Ding, Fan, Xu, and Yang}]{Ding2020}
\bibinfo{author}{Y.~Ding}, \bibinfo{author}{H.~Fan}, \bibinfo{author}{M.~Xu},
  \bibinfo{author}{Y.~Yang},
\newblock \bibinfo{title}{Adaptive exploration for unsupervised person
  re-identification},
\newblock \bibinfo{journal}{{ACM} Transactions on Multimedia Computing,
  Communications, and Applications} \bibinfo{volume}{16} (\bibinfo{year}{2020})
  \bibinfo{pages}{1--19}. \URLprefix \url{https://doi.org/10.1145/3369393}.
  \DOIprefix\doi{10.1145/3369393}.

\end{thebibliography}







\end{document}